\pdfoutput=1

\documentclass[11pt]{article}

\usepackage[final]{acl}

\usepackage{times}
\usepackage{latexsym}

\usepackage[T1]{fontenc}

\usepackage[utf8]{inputenc}

\usepackage{microtype}

\usepackage{inconsolata}

\usepackage{graphicx}

\usepackage{booktabs}
\usepackage{array}

\usepackage{cleveref}

\usepackage{enumitem}
\usepackage{array} 
\usepackage{float}

\title{Who Wrote This? Identifying Machine vs Human-Generated Text in Hausa}

\author{
    Babangida Sani$^{1}$, Aakansha Soy$^{1}$, Sukairaj Hafiz Imam$^{2}$, Ahmad Mustapha$^{1}$,\\
    \bf Lukman Jibril Aliyu$^{3}$, Idris Abdulmumin$^{4}$, Ibrahim Said Ahmad$^{5}$,\\ 
    \bf Shamsuddeen Hassan Muhammad$^{6}$\\ 
    \footnotesize $^1$Kalinga University, $^2$Bayero University, Kano,
    $^3$Arewa Data Science Academy, \\
    \footnotesize  $^4$DSFSI, University of Pretoria, $^5$Northeastern University, $^6$Imperial College London\\ 
    \footnotesize \texttt{\textbf{correspondence}: bsani488@gmail.com}
}

\begin{document}
\maketitle

\begin{abstract}
The advancement of large language models (LLMs) has allowed them to be proficient in various tasks, including content generation. However, their unregulated usage can lead to malicious activities such as plagiarism and generating and spreading fake news, especially for low-resource languages. 
Most existing machine-generated text detectors are trained on high-resource languages like English, French, etc. In this study, we developed the first large-scale detector that can distinguish between human- and machine-generated content in Hausa. We scrapped seven Hausa-language media outlets for the human-generated text and 
the Gemini-2.0 flash model to automatically generate the corresponding Hausa-language articles based on the human-generated article headlines.
We fine-tuned four pre-trained Afri-centric models (AfriTeVa, AfriBERTa, AfroXLMR, and AfroXLMR-76L) on the resulting dataset and assessed their performance using accuracy and F1-score metrics. AfroXLMR achieved the highest performance with an accuracy of 99.23\% and an F1 score of 99.21\%, demonstrating its effectiveness for Hausa text detection. Our dataset is made publicly available\footnote{\url{https://github.com/TheBangis/hausa_corpus}} to enable further research.
\end{abstract}

\section{Introduction}

Hausa is among the most spoken Chadic languages, belonging to the Afroasiatic phylum. Over 100 million people are estimated to speak the language, with the majority of speakers living in Northern Nigeria and the Republic of Niger, respectively \cite{r1inuwa2021first}. However, from computational linguistics, it is regarded as a low-resource language, having insufficient resources to support tasks involving Natural Language Processing (NLP; \citealt{r2adam2023detection,muhammad-etal-2023-afrisenti}).

Hausa language is written in either the Latin (or \textit{\textbf{Boko}}) and Arabic (or \textit{\textbf{Ajami}}) script \cite{Jaggar2006}. The \textit{Boko} script, existing since the 1930s, was introduced by the British colonial administration, and is used in education, government, and digital communication. The \textit{Ajami} script, an order writing system of the Hausa language that existed in pre-colonial times, is used mostly in religious, cultural, and informal writing. For the purpose of our work, and as Hausa is widely written nowadays, we scraped and generated data based on the Latin-based script.

Large language models (LLMs) are becoming mainstream and easily accessible, ushering in an explosion of machine-generated content over various channels, such as news, social media, question-answering (QA) forums, educational, and even academic contexts \cite{r3wang2023m4}. The human-like quality of texts generated by LLMs models for different languages including Hausa language is always advancing, allowing them to generate diverse content. LLMs, intentionally or unintentionally, have the potential to be used to create and propagate harmful or misleading content, such as fake news or hate speech \cite{r4xie2024mugc}, or even fake or artificial scholarship. To ensure the authenticity, accuracy, and trustworthiness of content, there is a need for machine-generated text detectors. Extensive research has been undertaken to differentiate between machine-generated texts (MGTs) and human-generated texts (HGTs), primarily employing model-based approaches \cite{r3wang2023m4, r5alshammari2024ai, r6ji2024detecting}.

In existing studies, (i) focus has mainly been on high-resource languages like English; (ii) there are no reliable detectors for detecting human vs. AI-generated text in the Hausa language; (iii) ensuring content authenticity is difficult, especially for low resource languages like Hausa \cite{r6ji2024detecting}. We aim, therefore, to develop an automatic detector to classify human-generated and machine-generated text in Hausa, focused on the news domain, hence filling this gap. The following are our contributions: 

\begin{itemize}
    \item We are the first to develop a Hausa detector that is capable of differentiating HGT and MGT in Hausa. We believe it would help in ensuring content authenticity in digital communication, academia, and mitigating fake news. 
    
    \item We curated a dataset that consists of human-generated data by scraping seven Hausa media outlets and machine-generated data using Gemini, addressing the lack of high-quality data in the area.
    
    \item By focusing on the Hausa language, we contribute to the expanding NLP capabilities for low-resource languages.
    

    \item All our resources will be open-source to encourage future academic research in the Hausa language.
    
\end{itemize}

\section{Related Work}
\subsection{Detection of MGTs before ChatGPT}
\citet{r9radford2019language} raised concerns regarding using machine-generated text for malicious purposes such as spam, fake news, plagiarism, and disinformation. The GLTR (Giant Language Model Test Room) tool \cite{r7gehrmann2019gltr}, released in June 2019, is an open-source system for detecting GPT-2-generated text using baseline statistical methods. Later that year, OpenAI enhanced the Roberta model \cite{r8liu2019roberta} by introducing a dedicated GPT-2 detector \cite{r9radford2019language}. Another major advancement was the GROVER model \cite{r10zellers2019defending}, which can both generate and detect fake news. With 5,000 self-generated articles and extensive real news content, GROVER achieved a 92\% detection accuracy, surpassing models like the Plug and Play Language Model (PPLM) \cite{r11dathathri2019plug} and BERT \cite{r12devlin2019bert}. Another study by \citet{r13ippolito2019automatic} examines the detection of machine-generated texts (MGTs) from GPT-2 with decoding strategies such as top-k, untruncated random sampling, and nucleus sampling in English. They discovered that optimized BERT was best but had poor cross-strategy generalization, whereas automatic classifiers performed better than humans, who misclassified AI text more than 30\% of the time. AraGPT-2 \cite{r14antoun2020aragpt2} introduced the first advanced Arabic language model that aides in distinguishing human-written and machine-generated Arabic text.

\subsection{Detection of MGTs after ChatGPT}
The launch of ChatGPT in late 2022 and later sequential models like GPT-4, Gemini, Claude, Llama, DeepSeek, etc., have posed new challenges as machine-generated texts (MGTs) mimic human writing styles more effectively than ever before. This raises concerns and the need for detection models to discern between HGTs and MGTs in different fields, such as academia, to mitigate plagiarism. In 2024, a study by \citet{r16jawaid2024human} presents a systemic approach for discerning between HWTs and MGTs using a combination of deep learning models, textual feature-based models, and machine learning models. Similarly, \citet{r4xie2024mugc} used eight traditional machine learning models and integrate statistical analysis, linguistic patterns, sentiment analysis and fact-checking as factors to differentiate between human-generated and machine-generated content across the three different datasets. In another study, \citet{r17mitrovic2023chatgpt} examines ChatGPT-generated short text detection using DistilBERT and a perplexity-based classifier on online reviews, creating three datasets: human-written, ChatGPT-generated, and ChatGPT-rephrased. DistilBERT achieved 98\% accuracy on original AI-generated text but only 79\% on rephrased text, indicating the challenge of detecting AI-rewritten text.

\section{Methodology}

\subsection{Datasets}
We used both human-generated text (HGTs) and machine-generated text (MTGTs) for the Hausa language in the news domain. We collected 2,586 HGTs from seven different local and international news outlets and generated equal amounts of texts for the MGTs by leveraging the Gemini-2.0-flash closed-source model. We merged the datasets into a single file and created a source column to label whether a text is a HGT or MGT and then shuffled the data for effective training and evaluation.
\Cref{tab:data} provides information on the composition of our dataset in terms of the number of sentences, words, and unique words.

\begin{table}[tbp]
    \centering
    \begin{tabular}{p{4cm}r}
        \toprule
        \textbf{Statistic} & \textbf{Count} \\
        \midrule
        Total Sentences & 6,737 \\
        Total Words & 3,376,976 \\
        Unique Words & 49,883 \\
        \bottomrule
    \end{tabular}
    \caption{\label{tab:data} Statistics of the dataset used in Hausa Machine-generated text detection.}
\end{table}



\begin{figure}[t!]
    \centering
    \includegraphics[width=\linewidth]{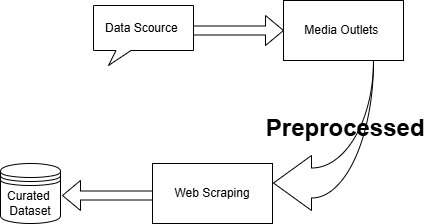}
    \caption{Overview of the pipeline's data collection for human-generated texts.}
    \label{fig:data_processing}
\end{figure}

\paragraph{Human-Generated Data}
The human-generated data was extracted from seven different local and international online news outlet websites written in the Hausa language through web scraping, structured into headlines and content. Initially, we extracted 3,700 news articles; and after automatically filtering out unwanted texts, we preprocessed the dataset to remove rows with empty content, reducing the news articles to exactly 2,586. \Cref{fig:data_processing} shows an overview of the pipeline's data collection process for human-generated data. 

\paragraph{Machine-Generated Data}
For the machine-generated text, we used the Gemini-2.0 flash model, through Google AI Studio, to automatically generate corresponding Hausa news articles based on each of the article headlines in the human-generated text. The dataset consists of headlines, content, and a word count for every article so that the text length of the generated article is near or equal to the actual article. To produce the machine-generated articles, we processed the dataset in batches of 10 and checked whether machine-generated text existed and initiated missing values where required. For every batch, the model generated full articles from the headlines, and after every batch, progress was saved so as not to lose data, and a 10-second delay was added between batches to avoid exceeding API limits. This iterative process continued until machine-generated articles were created for all headlines.

\section*{Data Processing}
The data collected, particularly the human-generated data, was noisy, containing many duplicates, English content, and other markup language symbols. Furthermore, some rows in the generated data produced an error message, while others included headlines that required cleaning. We identified and removed all unwanted content, including URLs, in both datasets before merging them into a single file to ensure effective training of our detection model.

\section{Experiments setup}
For our experiments, we utilized four Afri-centric transformer pre-trained language models: three multilingual and one monolingual. These models were selected due to their prior optimization for African languages, including Hausa. The models are AfriTeVa \cite{jude-ogundepo-etal-2022-afriteva}, AfroXLMR-76L \cite{r23adelani2023sib}, AfroXLMR \cite{alabi-etal-2022-adapting}, and AfriBERTa \cite{ogueji-etal-2021-small}, respectively. Each pre-trained language model was fine-tuned on our constructed dataset by training it up 3 epochs. The models were optimized with AdamW, with a learning rate of 1e-5, a batch size of 8, and a maximum sequence length of 512, with evaluation performed after each epoch. Table 2 shows the combinations of hyperparameters used to train the four models. The experiments were performed using PyTorch and Hugging Face Transformers. Upon completion, each fine-tuned model and tokenizer was saved and pushed to the Hugging Face Hub.

\begin{table}[t!]
\centering
\begin{tabular}{p{4cm}r}
\hline
\textbf{Hyperparameter} & \textbf{Value}  \\
\hline
optimizer       & AdamW   \\
epochs          & 3       \\
batch size      & 8       \\
learning rate   & 1e-5    \\
\hline
\end{tabular}
\caption{\label{tab:data} Hyperparameters used for training the pre-trained language models.}
\end{table}

\section{Results and Discussion}

\subsection{Results}

\Cref{tab:data} shows the performance of the fine-tuned models. The models' good performances indicate their capabilities to
distinguish between machine- and human-generated news texts written in Hausa language. Consistent in many downstream tasks, AfroXLMR performed the best, with an accuracy of 0.9923 and an F1 score of 0.9921. This is followed by AfriTeVa and AfriBERTa with an accuracy of 0.9884 and 0.9807 and an F1 score of 0.9881 and 0.9805, respectively, while AfroXLMR-76L had the lowest performance with an accuracy of 0.9672 and an F1 score of 0.9674. However, the overall performances indicate that all the developed models are very capable of detecting texts that are automaticall generated from human-written news articles.

\subsection{Discussion}
The results of our experiments reveal the efficacy of pre-trained language models in detecting between human-generated text (HGTs) and machine-generated text (MGTs) in the Hausa language. AfroXLMR, was the best-performing model, achieved an accuracy of 99.23\% on the test set and an F1 score of 99.21\%, indicating its efficacy in identifying text origins with minimal misclassification. This suggests that multilingual pre-trained language models optimized for African languages can be fine-tuned effectively for low-resource language tasks such as MGT detection. Relative to the other three models, AfriTeVa, AfriBERTa, and AfroXLMR-76L showed different levels of performance.  AfriTeVa achieved an accuracy of 98.84\% on the test set, followed by AfriBERTa with 98.07\% accuracy on the test set and lastly the AfroXLMR-76L achieved the lowest accuracy of 96.72\% on the test set. This difference can be due to model architecture, pretraining data, and optimization methods.

\begin{table}[t!]
    \centering
    \begin{tabular}{lcc}
        \toprule
        \textbf{Model} & \textbf{Accuracy} & \textbf{F1 Score} \\
        \midrule
        AfriTeVa & 0.9884 & 0.9881 \\
        AfriBERTa & 0.9807 & 0.9805 \\ 
        \textbf{AfroXLMR} & \textbf{0.9923} & \textbf{0.9921} \\
        AfroXLMR-76L & 0.9672 & 0.9674 \\
        \bottomrule
    \end{tabular}
    \caption{\label{tab:data} Results and performance of the fine-tuned models. The best-performing model is highlighted in bold.}
\end{table}

\section{Conclusion and Future Work}

In this paper, we introduced the first large-scale effort to develop a detector capable of distinguishing between human-generated text (HGT) and machine-generated text (MGT) in the Hausa language. The study consists of two main parts. Firstly, we created a dataset consisting of both human-generated and machine-generated. Next, we developed and evaluated the detectors by fine-tuning four Afri-centric pre-trained language models on the dataset. The models are AfriTeVa, AfriBERTa, AfroXLMR, and AfroXLMR-76L. We trained the models multiple times to optimize hyperparameters and enhance performance. The experimental results revealed the efficacy of the proposed models, with AfroXLMR outperforming the other models, achieving an accuracy of 99.23\% and an F1 score of 99.21\%. 

This study not only advances the detection of human-generated text and machine-generated text in a low-resource language such as Hausa but also shows that multilingual models optimized for African languages can be effectively adapted for detecting machine-generated text in low-resource languages. Support for low-resource languages is continuously improving across various large language models (LLMs). As a result, effective detection is important to prevent the spread of misinformation and disinformation, which are often facilitated by these models. We anticipate that this study will offer a comprehensive assessment of detection capabilities and enhance the ongoing academic discourse on identifying content generated by language models especially in underserved languages.

For future research, we aim to extend the dataset to cover diverse domains beyond news articles such as social media posts, academic writing, and books, as well as increasing the dataset size for better model generalization. Secondly, we aim to create real-time detection frameworks for use on digital platforms to help mitigate the propagation of AI-driven misinformation. Thirdly, exploring the use of the GPTs and other large language models in identifying machine-generated Hausa text. Using these models, with their high-level knowledge of language and context, detection accuracy could be enhanced. Lastly, to expand detection capabilities to other low-resource African languages, future research might explore cross-language transfer learning.

\section{Limitations}

Our study also has some limitations. First, we focused only on one domain when creating our dataset, which is news articles.
The training was limited to three epochs, and a small batch size of 8 across all the models, which may impact the models' performance. Lastly, the machine-generated data was sourced only from the Gemini-2.0 flash model, which reduces the variety of machine-generated text the models can detect. 

\section*{Acknowledgments}

We thank HausaNLP Community for generously providing access to the Google Colab GPU Premium Version, greatly enhancing our model's training efficiency and supporting this research.

\bibliography{anthology,custom}




\end{document}